\documentclass[11pt]{article}

\usepackage[T1]{fontenc}
\usepackage[utf8]{inputenc}   
\usepackage[english]{babel}

\usepackage{graphicx}   
\usepackage{booktabs}   
\usepackage{xcolor}

\usepackage{hyperref}  

\usepackage{xspace}

\usepackage{caption}   
\usepackage{makecell}  
\usepackage{pifont}
\usepackage{amsmath}
\usepackage{amssymb}
\usepackage{xspace}
\usepackage{enumitem}
\usepackage{authblk}

\usepackage[margin=1in]{geometry}  

\title{Seeing Through Reflections: Advancing 3D Scene Reconstruction \\ in Mirror-Containing Environments  with Gaussian Splatting}

\author[1]{Zijing Guo}
\author[1]{Yunyang Zhao}
\author[2]{Lin Wang}

\affil[1]{School of Automation and Intelligent Sening, Shanghai Jiao Tong University}
\affil[2]{Ningbo Artificial Intelligence Institute, Shanghai Jiao Tong University}
\date{}

\begin{document}

\maketitle

\begin{abstract}
   Mirror-containing environments pose unique challenges for 3D reconstruction and novel view synthesis (NVS), as reflective surfaces introduce view-dependent distortions and inconsistencies. While cutting-edge methods such as Neural Radiance Fields (NeRF) and 3D Gaussian Splatting (3DGS) excel in typical scenes, their performance deteriorates in the presence of mirrors. Existing solutions mainly focus on handling mirror surfaces through symmetry mapping but often overlook the rich information carried by mirror reflections. These reflections offer complementary perspectives that can fill in absent details and significantly enhance reconstruction quality. To advance 3D reconstruction in mirror-rich environments, we present \textbf{MirrorScene3D}, a comprehensive dataset featuring diverse indoor scenes, 1256 high-quality images, and annotated mirror masks, providing a benchmark for evaluating reconstruction methods in reflective settings. Building on this, we propose \textbf{ReflectiveGS}, an extension of 3D Gaussian Splatting that utilizes mirror reflections as complementary viewpoints rather than simple symmetry artifacts, enhancing scene geometry and recovering absent details. Experiments on MirrorScene3D show that ReflectiveGaussian outperforms existing methods in SSIM, PSNR, LPIPS, and training speed, setting a new benchmark for 3D reconstruction in mirror-rich environments.

\end{abstract}  

\section{Introduction}
\label{sec:intro}

Mirror-containing scenes are ubiquitous in modern environments, from homes with reflective furniture to buildings with glass facades. These reflective surfaces, while visually striking, pose significant challenges for 3D reconstruction and novel view synthesis (NVS), both of which are critical for applications in virtual reality, robotics, and computer graphics. Unlike diffuse surfaces, mirrors introduce view-dependent distortions and occlusions, making it difficult for traditional reconstruction pipelines to accurately capture and render such scenes \cite{jiang2024gaussianshader,liu2023nero,yang2025spec,yin2023multi}.

In recent years, Neural Radiance Fields (NeRF) \cite{mildenhall2021nerf} and 3D Gaussian Splatting (3DGS) \cite{kerbl20233d} have revolutionized 3D reconstruction. NeRF models density and view-dependent colors of scene points using neural implicit fields, while 3DGS represents scenes explicitly using Gaussian ellipsoids with anisotropic covariance matrices. These methods achieve remarkable performance in typical scenes \cite{zhang2020nerf++,wang2021neus,fridovich2022plenoxels}, with 3DGS further excelling in real-time rendering and visual fidelity \cite{zwicker2001ewa,yan2024multi,liang2024analytic}. However, their effectiveness diminishes in mirror-containing environments, where view-inconsistent reflections disrupt reconstruction pipelines. Existing methods, such as Mirror-NeRF \cite{zeng2023mirror} and TraM-NeRF \cite{holland2024tram}, address this issue by focusing on rendering the mirror surface itself, using techniques like ray tracing and symmetry mapping to generate reflections. Similarly, 3DGS-based approaches, including MirrorGaussian \cite{liu2024mirrorgaussian}, have improved visual quality by blending mirror and non-mirror regions. While effective for mirror rendering, these methods neglect a critical aspect: the valuable information embedded in mirror reflections. Reflected content offers complementary viewpoints that can fill in absent details of the real scene, a potential that has been rarely explored in previous work.

\begin{figure}[t]
    \centering
    \includegraphics[width=0.7\linewidth]{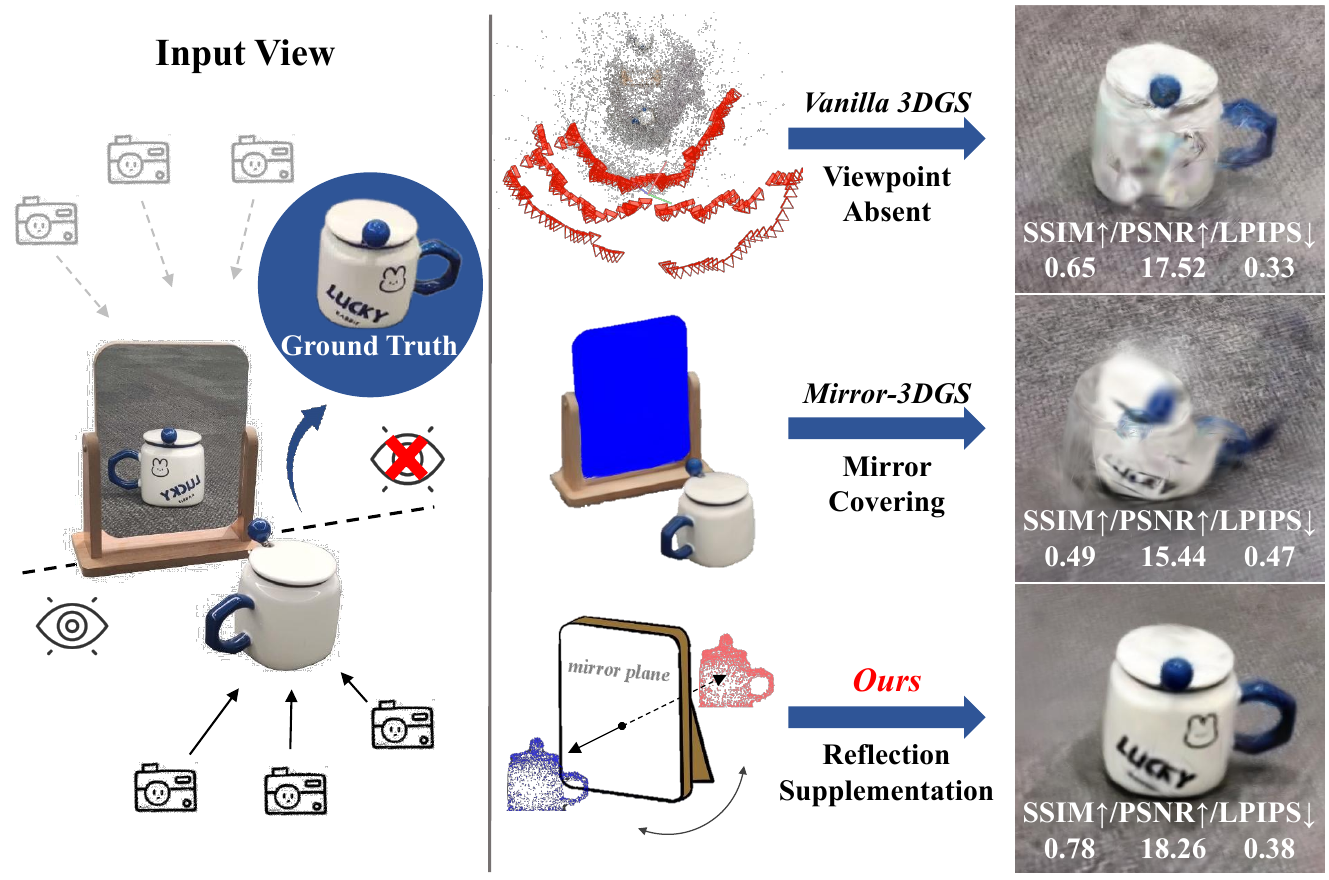}
    \caption{\textbf{Comparison of Methods and Rendering Performance of Vanilla3DGS, Mirror3DGS, and ReflectiveGS (Ours) on the ‘Cup’ Scene}. Vanilla 3DGS is unable to capture viewpoints on the mirror side, leading to incomplete reconstructions. Mirror-3DGS focuses on refining mirror surface details by distinguishing mirror and non-mirror regions using a blue mask. While this prevents interference from virtual 3D Gaussians, it limits the method’s ability to utilize reflection-based information for reconstruction. ReflectiveGS (Ours) incorporates supplementary mirror data to recover occluded details, enhancing reconstruction quality by integrating mirror reflections into joint training. Compared to other methods, our approach achieves more realistic and detailed renderings while maintaining competitive training times.
}
    \label{fig:introduction}
\end{figure}

To address this gap, we introduce MirrorScene3D, a new dataset specifically designed for mirror-augmented reconstruction (Using information contained in mirror to accomplish object-cnetric 3D Reconstruction, One example in Figure~\ref{fig:introduction}). It contains five diverse scenes with approximately 1,000 images and corresponding mirror masks (see Table~\ref{tab:dataset_comparison}), providing a benchmark for evaluating reconstruction methods in reflective environments.

Building on this, we propose \textbf{ReflectiveGS}, a novel extension of 3D Gaussian Splatting that explicitly integrates mirror reflection data into the reconstruction process. Rather than discarding reflections or treating mirrors as simple symmetry artifacts, ReflectiveGS leverages reflected content as additional viewpoints, enriching scene geometry and filling in absent details. Our pipeline incorporates multi-view object-centric images with mirror masks and introduces a symmetry consistency loss to jointly optimize real and reflected objects, ensuring structural coherence. Our method achieves improvements across all metrics compared to the baseline methods, with an average 30\% SSIM increase, 2.0 dB PSNR boost, and 20\% LPIPS reduction, while maintaining competitive training time, balancing visual quality and computational efficiency.

\begin{table*}[t]
\centering
\resizebox{\textwidth}{!}{ 
\begin{tabular}{@{}l|llccc@{}}
\toprule
\textbf{Dataset}    & \textbf{Task Focus}            & \textbf{Scene Type}      & \textbf{Scene Numbers}      & \textbf{Avg. Images}       & \textbf{Mirror Info} \\ \midrule
MSD \cite{yang2019my}       & Mirror Segmentation        & Independent Images       & -                           & -                         & \ding{55}                   \\
PMD \cite{lin2020progressive}   & Mirror Segmentation        & Independent Images       & -                           & -                         &    \ding{55}                \\
Mirror-NeRF\cite{zeng2023mirror}       & Mirror Rendering           & Complete Scenes          & 5 Synthetic, 4 Real         & 200\textasciitilde{}300   &   \ding{55}                 \\
\textbf{MirrorScene3D(Ours)}     & Mirror Supplementation     & Complete Scenes          & 5 Real                      & 200\textasciitilde{}300   & \checkmark                   \\ \bottomrule
\end{tabular}
}
 \caption{\textbf{Comparison of Our Benchmark with Existing Datasets.} Our dataset (MirrorScene3D) is specifically designed to validate algorithms for 3D scene reconstruction that utilize mirror information and address the supplementation of absent details. It includes five comprehensive real-world scenes, with an average scale comparable to previous datasets, while also integrating exploitable mirror data for augmented reconstruction accuracy.}
\label{tab:dataset_comparison}
\end{table*}

In summary, our contributions are as follows:

\begin{itemize}

\item We introduce \textbf{MirrorScene3D}, a novel dataset designed for mirror-augmented 3D reconstruction, offering high-quality multi-view images and mirror masks to benchmark reconstruction methods in reflective environments.

\item We propose \textbf{ReflectiveGS}, an approach that fully integrates mirror reflection data into 3D reconstruction, leveraging reflections as complementary viewpoints to enhance reconstruction quality in mirror-rich scenes.

\item We develop a novel optimization strategy based on symmetry consistency loss, enabling the joint training of real and reflected objects to ensure structural coherence and detail preservation.

\end{itemize}

\section{Related Works}
\label{related}

This section reviews two categories of work closely related to our research: novel view synthesis and mirror-aware reconstruction.
\subsection{Novel View Synthesis}
NVS (Novel View Synthesis) generates new images from novel viewpoints using a limited set of known perspectives~\cite{chan2023generative, wiles2020synsin, mildenhall2019local}. By learning the structure of objects or scenes from multiple angles, it synthesizes images from new viewpoints. The introduction of NeRF (Neural Radiance Fields) in 2020 \cite{mildenhall2021nerf} marked a significant breakthrough. NeRF uses MLPs to estimate density and view-dependent colors for points via ray marching and volume rendering, producing impressive visuals. Research on NeRF has advanced rapidly, focusing on speed, quality, and handling challenging scenes \cite{muller2022instant, yu2021plenoctrees, deng2022depth, barron2023zip, barron2021mip, yu2021pixelnerf, martin2021nerf, pumarola2021d}. However, its reliance on dense sampling and computationally intensive ray marching remains a challenge, particularly in real-time applications or incomplete viewpoints.

The emergence of 3DGS (3D Gaussian Splatting) \cite{kerbl20233d} in 2023 has further advanced the development of this field~\cite{
wu20244d, yu2024mip, yang2024deformable, lee2024compact, tang2023dreamgaussian, huang20242d}. In contrast to NeRF, 3DGS explicitly represents 3D scenes using anisotropic Gaussians and leverages a tile-based rasterizer that allows $\alpha$-blending. This enables substantial improvements in training efficiency, visual quality, and real-time rendering, making it well-suited for interactive applications. Furthermore, 3DGS is applied in various domains \cite{wu2024recent}, such as scene understanding and segmentation \cite{shi2024language,ye2023gaussian,huang2024point,cen2023segment,zhou2024feature}, SLAM (Simultaneous Localization and Mapping) \cite{yan2024gs,matsuki2024gaussian,keetha2024splatam,yugay2023gaussian,hong2024liv}, and more. 

Despite its advantages, 3DGS struggles with novel view synthesis under incomplete viewpoints. To overcome this, we incorporate mirror reflection as supplementary information and apply mirror symmetry to reconstruct absent viewpoints within the 3DGS framework.

\subsection{Mirror-Aware Reconstruction}

Mirrors are a crucial source of supplementary information in our research. Existing studies on mirror scenes primarily focus on rendering mirror regions, where specular reflections cause multi-view inconsistencies, leading to blurriness or aliasing. NeRF-based methods have attempted to address these issues: Ref-NeRF \cite{verbin2022ref} reparameterizes directional MLPs for improved view-dependent results, while Mirror-NeRF \cite{zeng2023mirror} models reflection probability to blend camera and reflected rays. However, these methods suffer from long runtimes, limiting their efficiency and quality.

Recent 3DGS-based methods offer new perspectives. Mirror-3DGS \cite{meng2024mirror} introduces mirror attributes into 3DGS to derive mirror transformation, enhancing the realism of mirror rendering through two-stage training process. MirrorGaussian~\cite{liu2024mirrorgaussian} proposes a dual-rendering approach that enables differentiable rasterization of both the real-world 3D Gaussians and their mirrored counterparts.

However, these methods discard mirror content and rely solely on physical symmetry to generate mirror rendering. In contrast, our approach captures and leverages mirror reflections to supplement object reconstruction, achieving more detailed results and better visual quality.

\section{MirrorScene3D Construction}
\label{data}

\subsection{Data Construction}

We conduct \textbf{MirrorScene3D}, a dataset designed to leverage mirror reflections for supplementing absent details in scene reconstruction. Existing mirror-related datasets mainly focus on mirror segmentation (e.g., MSD~\cite{yang2019my}, PMD~\cite{lin2020progressive}) or rendering effects of mirror regions (e.g., Mirror-NeRF~\cite{zeng2023mirror}), but none address the challenge of utilizing mirror reflections to recover occluded viewpoints.

Our dataset consists of diverse indoor scenes for validating our method. It targets static, object-centric tasks, where a mirror is placed on one side of the object, and the camera captures $180^{\circ}$ multi-view images from the opposite side. This setup ensures that occluded object regions are only visible through mirror reflections. To enable quantitative evaluation, we also capture complete multi-view images after removing the mirrors, providing ground truth for previously occluded areas. Table~\ref{tab:dataset_comparison} presents a comparison with related datasets.

\begin{figure}[t]
    \centering
    \includegraphics[width=0.7\linewidth]{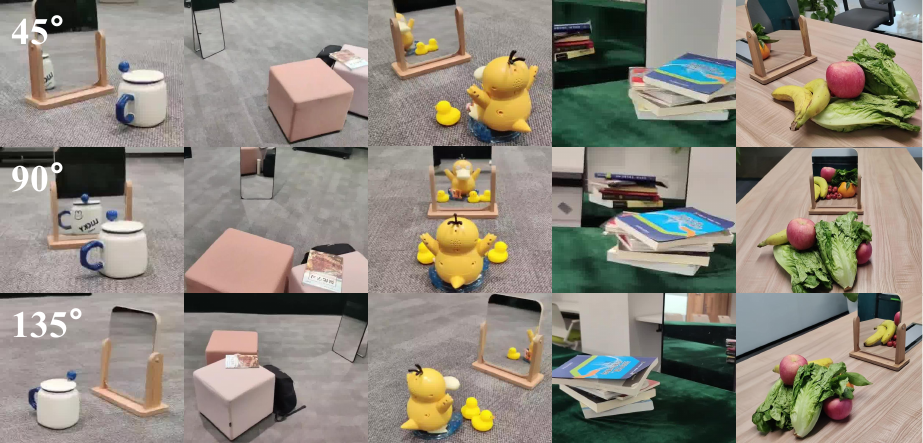}
    \caption{\textbf{Scenes from Our Dataset.} Our dataset captures $0^\circ-180^\circ$ multi-view images from the opposite side of the mirror, with the central object as the focus. Occlusion regions are set, which are only visible through the mirror reflection and not directly captured.}
    \label{fig:dataset}
\end{figure}

\begin{figure*}
    \centering
    \includegraphics[width=1\linewidth]{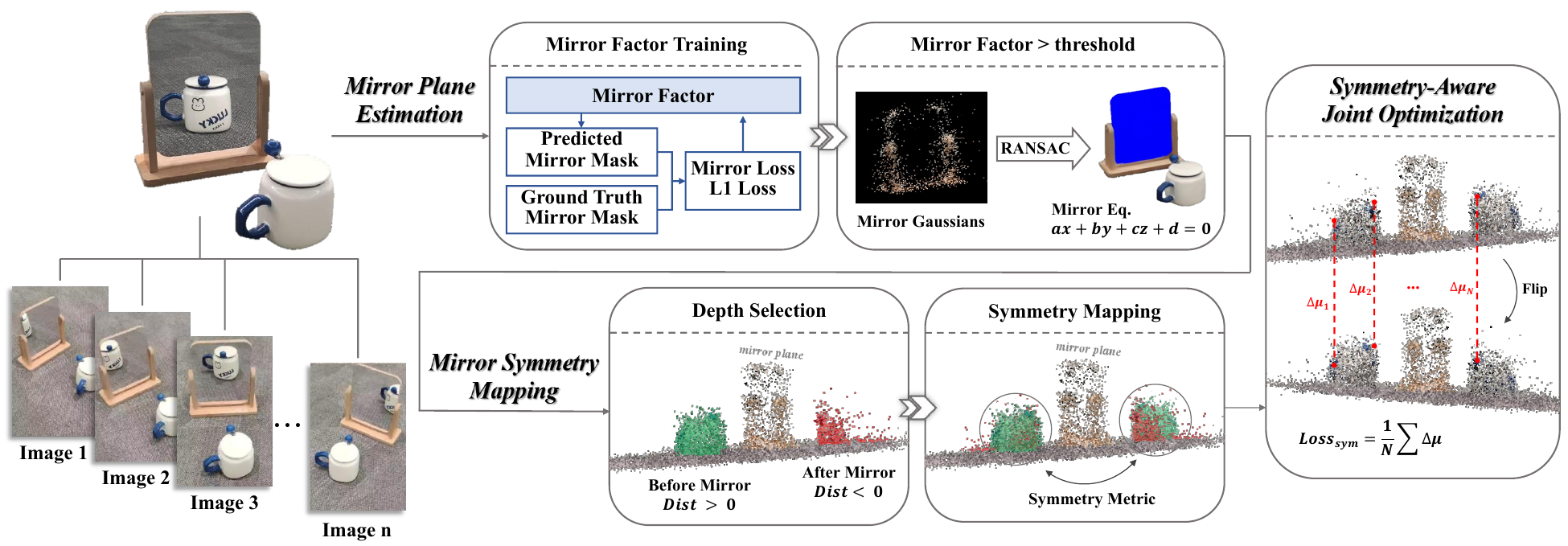}
    \caption{\textbf{Overview of the Method.} Our method takes images with viewpoint occlusions as input and utilizes three key steps to leverage mirror reflection information for augmented 3D scene reconstruction with more detail. First, we introduce a learnable mirror factor to represent the probability of mirror Gaussians, updating it via a loss function that compares the predicted mask with the ground truth. This allows us to fit the mirror equation and compute the corresponding symmetry metric. Next, based on conventional rendering results, we distinguish the Gaussians in front and behind the mirror by evaluating their distances to the mirror plane. The computed symmetry metric is then used to perform the symmetry mapping operation. To constrain training, we introduce a symmetry consistency loss, which minimizes the difference between the center coordinates of original and flipped Gaussians, ensuring collaborative optimization of Gaussians on both sides of the mirror.}
    \label{fig:pipeline}
\end{figure*}

\subsection{Scene Arrangement and Data Collection}

Our dataset includes diverse indoor scenes: residential settings (e.g., a cup, a sofa), toy scenes (e.g., a Psyduck figure, small ducks), office-like environments (e.g., a stack of books), and kitchen scenes (e.g., apples, bananas, leafy greens). The objects span various categories, with geometric diversity ranging from regular cuboidal shapes (e.g., sofas, books) to irregular surfaces (e.g., toys, fruits), providing a robust benchmark for reconstruction evaluation.

\noindent\textbf{Viewpoint Occlusion.} To define occluded regions, we designate specific areas in each scene: the front pattern of the cup (‘Cup’ scene), the bottle between bag and sofa (‘Sofa’ scene), musical notes under Psyduck (‘Duck’ scene), text and patterns on book spines (‘Book’ scene), and the hawthorn (‘Fruit’ scene). These areas remain unobserved in the $180^{\circ}$ multi-view capture but appear in mirror reflections. Figure~\ref{fig:dataset} illustrates the capture viewpoints. The reconstruction of these occluded regions serves as a key measure of algorithm’s ability to supplement absent details.

\noindent\textbf{Mirror Placement.} Mirror placement was standardized across all scenes to ensure consistency and comparability. Mirrors were positioned at a fixed height and distance from objects, maintaining uniform alignment and optimized angles to capture both objects and their surroundings without overlap or distortion. Calibration markers ensured precise positioning, and adjustments were made to ensure reflective regions covered key object features. This systematic setup provides reliable reflection data for evaluating reconstruction algorithms.

\noindent\textbf{Lighting Conditions.} Lighting conditions were carefully controlled to maintain consistency while accounting for environmental factors. A combination of natural and artificial light sources was used to replicate indoor settings, with optimized intensity and direction to minimize shadows, glare, and unwanted reflections. Considerations such as room layout, surface reflectivity, and light diffusion ensured uniform illumination across all scenes, enhancing the dataset’s reliability for reconstruction evaluation.

\subsection{Data Annotation and Preprocessing}

Mirror regions were manually annotated using LabelMe, saved as .json files, and converted to .jpg format for binary mask generation. All images were resized to $1280 \times 720$, resulting in a dataset of 1256 images across five diverse indoor scenes, serving as a benchmark for evaluating mirror-augmented reconstruction methods.

A standardized preprocessing pipeline was applied to ensure consistency. Gaussian filtering reduced image noise, while color correction addressed lighting variations. Images were resized for uniformity, and contrast adjustments enhanced visibility, especially in reflective regions, ensuring a well-structured dataset for reconstruction tasks.

\subsection{Summary}
\textbf{MirrorScene3D} is a new dataset designed for mirror-augmented 3D reconstruction, focusing on supplementing absent viewpoints through mirror reflections. It includes 1256 images spanning diverse indoor scenes with standardized mirror placement and lighting. The dataset provides a high-quality benchmark for assessing the integration of mirror reflections in reconstruction tasks.

\section{Method}
\label{method}

\noindent\textbf{Overview.} The method overview is illustrated in Figure~\ref{fig:pipeline}. ReflectiveGS leverages mirror information as supplementary data to improve the reconstruction of objects with absent views caused by occlusions in mirror-containing 3D scenes. The inputs to the method include multi-view images of the scene, corresponding mirror masks, and a sparse point cloud generated through structure-from-motion (SfM) \cite{schonberger2016structure}. While some object features are absent in direct views, they can often be observed in mirror reflections, providing valuable complementary information. Our method is structured around three key components: 

\begin{itemize}
    \item \textbf{Mirror Plane Estimation}: A parameterized mirror plane equation is used to establish the geometric relationship between real objects and their virtual reflections, enabling accurate feature mapping.  
    \item \textbf{Mirror Symmetry Mapping}: Leveraging the mirror plane equation, reflected objects are rendered through symmetry mapping. Depth selection ensures only valid reflected features are incorporated to complement real objects.  
    \item \textbf{Symmetry-Aware Joint Optimization}: A spatial symmetry consistency loss function is introduced to collaboratively optimize the real and reflected components of the point cloud, ensuring alignment and improving detail preservation.  
\end{itemize}  

Further details on these components are provided in Section~\ref{4.2} (Mirror Plane Estimation), Section~\ref{4.3} (Mirror Symmetry Mapping), and Section~\ref{4.4} (Symmetry-Aware Joint Optimization). Together, these steps enable ReflectiveGS to effectively utilize mirror reflections, enriching 3D scene reconstruction and addressing the limitations of traditional methods.

\subsection{Preliminaries}

3DGS is a method that explicitly represents 3D scenes with numerous anisotropic 3D Gaussian primitives. Each of them can be described as:
\begin{equation}
    \mathcal{G}_i(x) = exp\{-\frac{1}{2}(x - \mu_i)^T \Sigma_i^{-1} (x - \mu_i)\},
    \label{eq:gaussian_distribution}
\end{equation}
where $\mu_i \in \mathbb{R}^3$ represents the center position of $\mathcal{G}_i$, and $\Sigma_i \in \mathbb{R}^{3 \times 3}$ represents the 3D covariance matrix. 
$\Sigma_i$ also can be decomposed into a rotation matrix $R_i \in \mathbb{R}^{3 \times 3}$ and a scaling matrix $S_i \in \mathbb{R}^{3 \times 3}$, as shown in Equation~\ref{eq:sigma}.
\begin{equation}
    \Sigma_i=R_i S_i S_i^T R_i^T .
    \label{eq:sigma}
\end{equation}

The subsequent rasterization process projects the 3D Gaussian primitives onto the 2D screen space, with their shapes and extents determined by the covariance matrices. The final image is then generated by leveraging $\alpha$-Blending to integrate the colors and opacities of multiple Gaussians. Based on the differentiable rasterization principles of 3DGS, the rendering process is expressed as:
\begin{equation}
\begin{aligned}
    \textbf{C(p)} &= \sum_{i \in N} c_i(d) \alpha_i \mathcal{G}'_i(p) T_i, \\
    T_i &= \prod_{j=1}^{i-1} (1 - \alpha_j \mathcal{G}'_i(p), \\
    \mathcal{G}'_i(p) &= e^{ -\frac{1}{2} (p - \mu'_i)^T \Sigma_i'^{-1} (p - \mu'_i) },
\end{aligned}
\label{eq:render process}
\end{equation}
where $\textbf{C(p)}$ is the color of a pixel p, $\mathcal{G}'_i(p)$ represents the projected 2D Gaussians, $N$ represents the number of sample Gaussians, $c_i$ and $\alpha_i$ represent the color and opacity of the i-th Gaussian respectively.

\subsection{Mirror Plane Estimation}  
\label{4.2}

The foundation of 3D reconstruction in mirror-containing scenes lies in the principle of mirror symmetry. Accurately parameterizing the mirror plane is critical for linking real objects and their virtual counterparts. To achieve this, we introduce a learnable mirror factor \( m \in (0,1) \), an attribute of 3D Gaussians that represents the probability of a point belonging to the mirror. The mirror factor is combined with opacity \(\alpha\), yielding a modified opacity \(\alpha'_i = m_i \alpha_i\). The updated rendering equation is expressed as:  
\begin{equation}
    \textbf{C(p)} = \sum_{i \in N} c_i(d) \alpha'_i \mathcal{G}'_i(p) T_i, 
    \label{eq:render_new}
\end{equation}  
where \(\alpha_i\) denotes the probability of a ray intersecting the \(i\)-th Gaussian, and \(m_i\) adjusts visibility for mirror regions. High mirror factors enhance visibility for mirror regions, while non-mirror regions are rendered nearly transparent. The resulting rendered mirror mask is compared to the ground truth to refine the learning of the mirror factor.

Using the learned mirror factor, regions with high mirror probabilities are identified and used to fit a parameterized mirror plane equation via the RANSAC algorithm \cite{fischler1981random}. This mirror plane equation establishes the geometric relationship between real objects and their virtual counterparts.

\begin{figure}[t]
    \centering
    \includegraphics[width=0.7\linewidth]{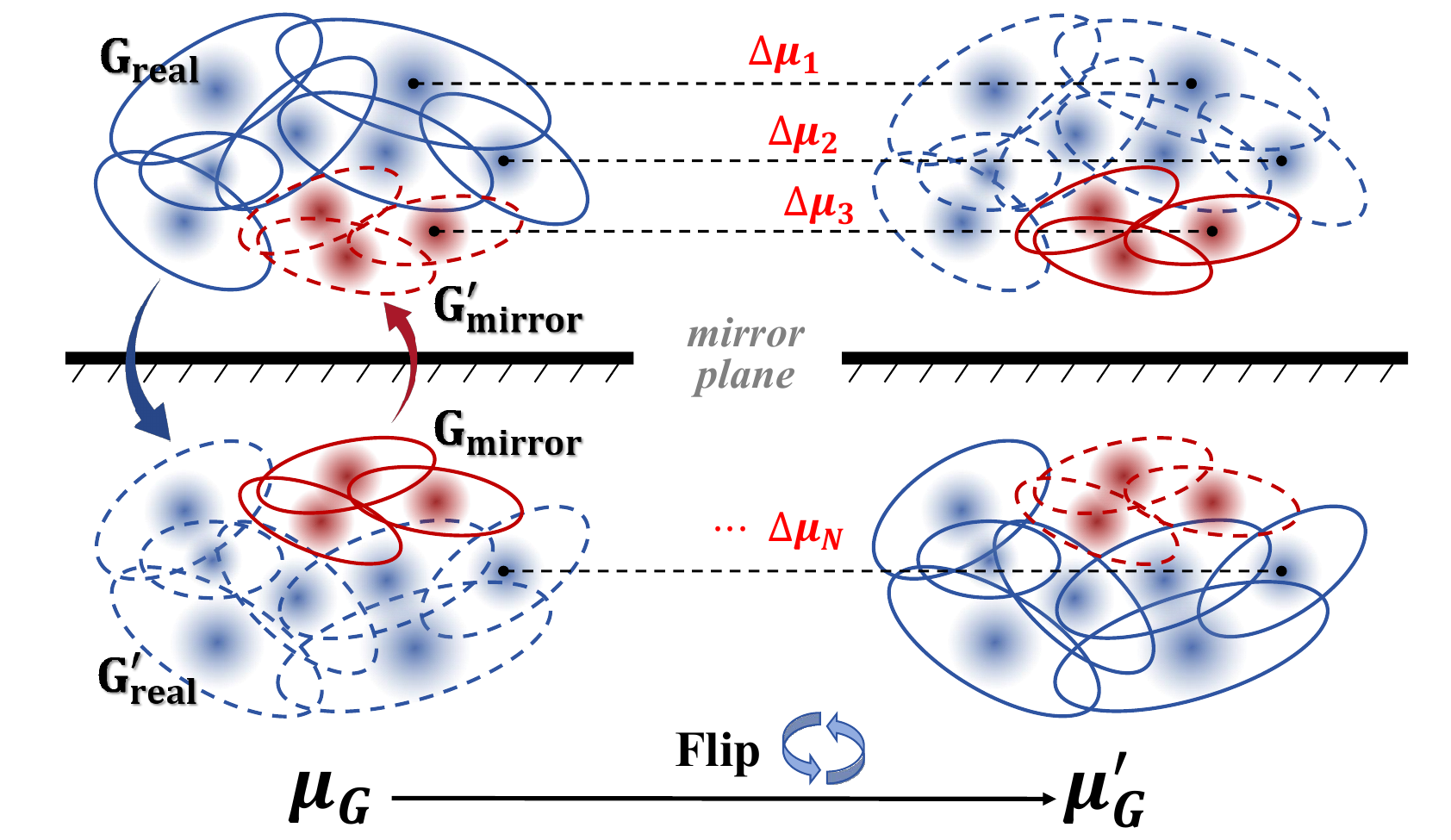}
    \caption{\textbf{The Illustration of Symmetry Consistency Loss Function.} In our method, we reflect the part directly captured ($G_{real}$ to $G'_{real}$), and merge it with the mirror part ($G_{mirror}$), using mirror region images to supervise training. We introduce a symmetry consistency loss, which measures the similarity between the $G_{real}$ and $G'_{real}$ to enable collaborative optimization. Since the number of Gaussians changes during training, we also symmetrically transform the $G_{mirror}$ to $G'_{mirror}$, and use the scene's overall symmetry to compare the center positions of the original Gaussians, computing the symmetry consistency loss.}
    \label{fig:loss}
\end{figure}

\subsection{Mirror Symmetry Mapping}  
\label{4.3}

Mirror symmetry is key to linking real 3D Gaussians with their virtual counterparts, enabling collaborative reconstruction. Symmetry mapping transforms 3D Gaussians across the mirror plane to align virtual counterparts with real objects. The transformation relies on the mirror plane equation \( ax + by + cz + d = 0 \), expressed as:
\begin{equation}
    \text{Symmetry} = \begin{bmatrix}
    1 - 2a^2 & -2ab & -2ac & -2ad \\
    -2ab & 1 - 2b^2 & -2bc & -2bd \\
    -2ac & -2bc & 1 - 2c^2 & -2cd \\
    0 & 0 & 0 & 1
    \end{bmatrix}.
    \label{eq:symmetry}
\end{equation}

For a Gaussian centered at \((x, y, z)\), the mirrored coordinates \((x', y', z')\) are derived as:
\begin{equation}
    (x', y', z', 1)^T = \text{Symmetry} \cdot (x, y, z, 1)^T.
    \label{eq:mirror_transform}
\end{equation}

Additionally, the symmetry of the rotation and view-dependent color attributes is adjusted to align the real and virtual components. Symmetrized Gaussians are then integrated to achieve a unified representation, providing complementary information for reconstruction.

\subsection{Symmetry-Aware Joint Optimization} 
\label{4.4}

To fully utilize mirror reflections, we introduce a spatial symmetry consistency loss that optimizes the integration of real and mirrored components. This loss measures the similarity between corresponding 3D Gaussians before and after symmetry transformation. As Shown in Figure~\ref{fig:loss}, the Gaussians' centers are compared using normalized mean coordinate differences. However, due to dynamic changes in the number of Gaussians during training, mirrored Gaussians are also reflected back to the front of the mirror for comparison. The symmetry consistency loss is formulated as:
\begin{equation}
    L_{\text{sym}} = \frac{1}{N} \sum \| \mu_G - \mu'_G \|,
    \label{eq:symmetry_loss}
\end{equation}
where $\mu_G = G_{real} + G'_{real} + G_{mirror} + G'_{mirror}$ and $\mu'_G$ represent the centers of flip Gaussians of $\mu_G$, and $N$is the total number of Gaussians in the scene. By minimizing this loss, the training process ensures alignment and consistent optimization of real and mirrored components, enhancing the overall reconstruction quality.

Based on the previous content, the composition of our complete loss function is as follows:
\begin{equation}
    L = L_{rgb} + \lambda_m L_m + \lambda_{sym} L_{sym},
\end{equation}
where $L_{rgb}$ is composed of the $L_1$ and D-SSIM loss, which supervises the learning of various attributes of the Gaussians. $L_m$ supervises the learning of the mirror factor, and $L_{sym}$ supervises the collaborative training of the specular front-and-back Gaussians. In our experiments, the hyperparameters are set to $\lambda_m = 1.0$ and $\lambda_{sym} = 10.0$.

\section{Experiments}
\label{experiments}

\begin{figure*}[t]
    \centering
    \includegraphics[width=1\linewidth]{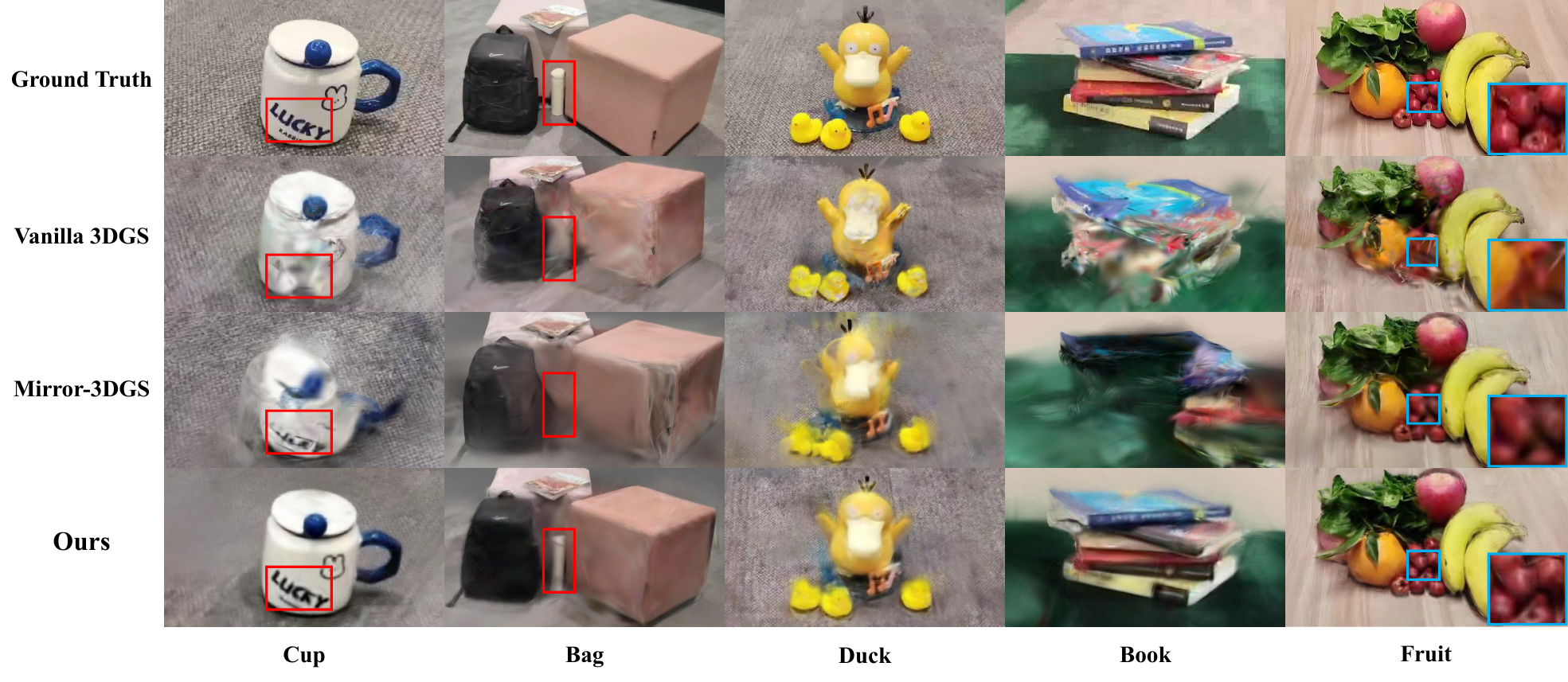}
    \caption{\textbf{Visualization of ReflectiveGS vs. Existing Methods Across Multiple Scenes in MirrorScene3D.} Column 1: Cup. Column 2: Bag. Column 3: Duck. Column 4: Book. Column 5: Fruit. From the visual results, our method yields rendering outcomes with clearer and richer details, demonstrating superior rendering quality.}
    \label{fig:compare}
\end{figure*}

\subsection{Experimental Setup}

\noindent\textbf{Metrics.}
We evaluate the performance using multiple metrics, including SSIM (Structural Similarity Index), PSNR (Peak Signal-to-Noise Ratio), LPIPS (Learned Perceptual Image Patch Similarity), and training time. These metrics collectively allow us to assess the similarity between rendered images and ground truth, providing a comprehensive measure of both visual fidelity and computational efficiency.
To conduct a comprehensive comparison, we perform a quantitative evaluation by resizing all images to a uniform resolution of 1280×720 across all methods. 

\noindent\textbf{Implementation.}
We performed experiments across all scenes in our dataset, evaluating Vanilla 3DGS, Mirror-3DGS, and our proposed ReflectiveGS method. The aforementioned metrics were recorded for each approach to allow for a detailed comparison. All experiments were executed on a system equipped with a GeForce RTX 3090 GPU, ensuring consistent hardware conditions for evaluating the methods.

\subsection{Experimental Results}

We conducted comprehensive experiments to evaluate our method, with visual results shown in Figure~\ref{fig:compare}. The following results demonstrate its advantages over baseline methods from multiple perspectives in mirror-augmented 3D reconstruction.

\begin{table*}[t]
\centering
\resizebox{\textwidth}{!}{ 
\begin{tabular}{l|cccc|cccc|cccc|cccc|cccc}
\toprule
\textbf{Scene}                 & \multicolumn{4}{c|}{\textit{Cup}}              & \multicolumn{4}{c|}{\textit{Bag}}              & \multicolumn{4}{c|}{\textit{Duck}}             & \multicolumn{4}{c|}{\textit{Book}}            & \multicolumn{4}{c}{\textit{Fruit}} \\ 
\cmidrule(lr){2-5} \cmidrule(lr){6-9} \cmidrule(lr){10-13} \cmidrule(lr){14-17} \cmidrule(lr){18-21}
\textbf{Metrics}               & SSIM$\uparrow$   & PSNR$\uparrow$   & LPIPS$\downarrow$   & \makecell{Avg. \\ Time} & SSIM$\uparrow$   & PSNR$\uparrow$   & LPIPS$\downarrow$   & \makecell{Avg. \\ Time} & SSIM$\uparrow$   & PSNR$\uparrow$   & LPIPS$\downarrow$   & \makecell{Avg. \\ Time} & SSIM$\uparrow$   & PSNR$\uparrow$   & LPIPS$\downarrow$   & \makecell{Avg. \\ Time} & SSIM$\uparrow$   & PSNR$\uparrow$   & LPIPS$\downarrow$   & \makecell{Avg. \\ Time} \\ 
\midrule
Vanilla 3DGS                   & 0.65    & 17.52   & \textbf{0.33} & 22m        & 0.59    & 15.07   & 0.42     & 12m        & 0.68    & 18.76   & \textbf{0.26} & 29m        & 0.65    & 13.92   & 0.44     & 13m     & 0.86    &18.58   & 0.22     & 18m   \\
Mirror-3DGS                 & 0.49    & 15.44   & 0.47     & 53m        & 0.74    & 15.97   & 0.36     & 43m        & 0.72    & 19.77   & 0.41     & 53m        & 0.45    & 10.70   & 0.54     & 41m       & 0.87    & 18.30   & 0.18     & 47m  \\
\textbf{Ours}                  & \textbf{0.78} & \textbf{18.26} & 0.38 & 53m        & \textbf{0.85} & \textbf{17.01} & \textbf{0.26} & 43m        & \textbf{0.76} & \textbf{20.07} & 0.33 & 58m        & \textbf{0.90} & \textbf{16.88} & \textbf{0.27} & 41m   & \textbf{0.90} & \textbf{19.57} & \textbf{0.15} & 55m     \\ 
\bottomrule
\end{tabular}
}
\caption{\textbf{Comparison of Performance.} We conducted experiments on the five scenes in our constructed dataset, evaluating Vanilla 3DGS, Mirror-3DGS, and ReflectiveGS (ours). The results were recorded for SSIM, PSNR, LPIPS, and training time, with the best values highlighted in bold. Our method achieves 23.8\% higher SSIM, 1.59 dB higher PSNR, and 13.3\% lower LPIPS compared to Vanilla 3DGS, with even greater improvements over Mirror-3DGS.}
\label{tab:metrics_comparison}
\end{table*}

\noindent\textbf{Mirror Estimation Error Analysis.}
Our method estimates mirror regions using learnable mirror factors, which are updated by comparing predicted masks with ground truth. While this effectively fits the mirror equation, some estimation errors persist. Experiments found that small errors have minimal impact on the results, and the resulting symmetry inaccuracies are optimized during the collaborative training process. When the mirror parameter error exceeds 5\%, noticeable ghosting of objects appears in the rendered images.

\begin{figure}[b]
    \centering
    \includegraphics[width=0.7\linewidth]{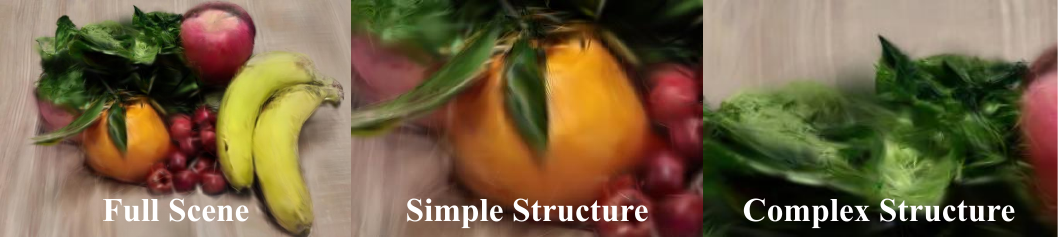}
    \caption{\textbf{Impact of Object Structure (‘Fruit’ Scene Example).} In this scene, simpler structures (e.g., orange) have clearer rendering results, while more complex structures (e.g., lettuce) show more blurred visual effects.}
    \label{fig:structure}
\end{figure}

\noindent\textbf{Impact of Object Structure.}
ReflectiveGS performs well on simple objects, effectively integrating reflected information with real object. However, its performance slightly declines on complex one. In Figure~\ref{fig:structure}, the geometry of objects affects the supplementation of mirrors. More complex surfaces require richer reflection data and higher precision in the mirror equation and symmetry consistency. However, our method maintains high rendering quality in general.

\noindent\textbf{Optimization and Training Efficiency.}
Vanilla 3DGS had the shortest training time (around 20 minutes), but its performance is limited in complex mirror-containing scenes. Although ReflectiveGS’s training time is longer (around 50 minutes), the improvements in mirror information utilization and absent detail supplementation justify the additional computational cost. Compared to baseline methods, our approach achieves superior rendering results while maintaining competitive training times.

\begin{figure}[t]
    \centering
    \includegraphics[width=0.7\linewidth]{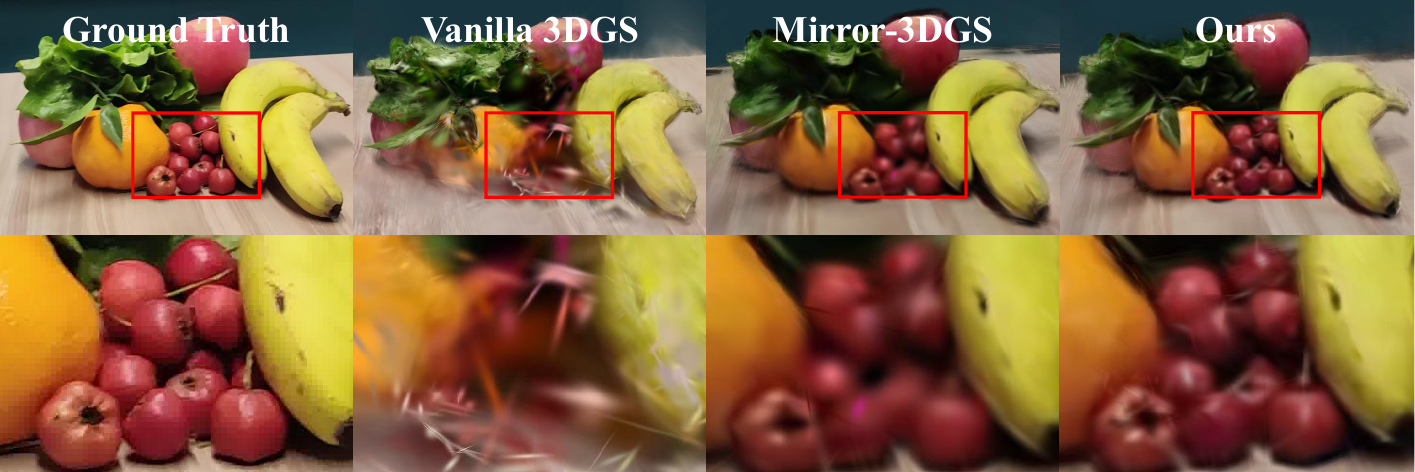}
    \caption{\textbf{Comparison of Full Scene (Row 1) and Detail View (Row 2)(‘Fruit’ Scene Example).} Our method utilizes mirror reflection information to enhance the 3D reconstruction, with the improvement being especially noticeable in the detail areas.}
    \label{fig:detail}
\end{figure}

\begin{table}[t]
\centering
\resizebox{0.7\columnwidth}{!}{ 
\begin{tabular}{l|ccc|ccc }
\toprule
\textbf{   }                 & \multicolumn{3}{c|}{\textit{Full Scene}}              & \multicolumn{3}{c}{\textit{Detail View}}  \\ 
\cmidrule(lr){2-4} \cmidrule(lr){5-7}
\textbf{Metrics}    & SSIM$\uparrow$   & PSNR$\uparrow$   & LPIPS$\downarrow$   & SSIM$\uparrow$   & PSNR$\uparrow$   & LPIPS$\downarrow$   \\
\midrule
Vanilla 3DGS                 & 0.84    & 16.79   &0.29   & 0.74    &15.1   & 0.54   \\
Mirror-3DGS              & 0.87 &16.61 & 0.21 & 0.87 & 16.51 & 0.45  \\
\textbf{ReflectiveGS}    & \textbf{0.90} & \textbf{17.67} &  \textbf{0.18}    & \textbf{0.91} & \textbf{17.72} & \textbf{0.36}       \\ 
\bottomrule
\end{tabular}
}
\caption{\textbf{Metrics of Full Scene and Detail View  (‘Fruit’ Scene Example).} Our method outperforms the baseline in all metrics, with the improvement being more pronounced in the detail view compared to the full scene, highlighting our method's advantage in supplementing occluded details.}
\label{tab:detail}
\end{table}

\noindent\textbf{Rendering Quality Analysis.}
We quantitatively evaluated rendering quality using SSIM, PSNR, and LPIPS (see Table~\ref{tab:metrics_comparison}). The results show that our method consistently outperforms the baseline methods in SSIM and PSNR, demonstrating its superiority in detail and accurate reconstruction. And it achieves the lowest LPIPS values in most scenes, indicating that it provides the most visually realistic results. From the analysis of specific scenes, our method shows the most significant improvement in the Cup and Book scenes, demonstrating its strength in reconstructing regular objects. It also performs well in the Bag and Fruit scenes, highlighting its advantage in detail supplementation. Additionally, in the Duck scene, our method achieved the best PSNR, proving its precision in color and brightness reconstruction.

\noindent\textbf{Detail Performance.}
To better assess the detail reconstruction, we conducted the same quantitative comparison using SSIM, PSNR and LPIPS for the occluded regions defined in Section~\ref{data}. The visual results (Figure~\ref{fig:detail}) and metrics (Table~\ref{tab:detail}) show a more significant improvement in detail view than full scene, with a higher improvement rate across all metrics relative to the baseline. This demonstrates the effectiveness of our method in utilizing mirror information for 3D reconstruction, particularly in supplementing details in viewpoint blind spots.

\noindent\textbf{Summary.}
Experimental results show that ReflectiveGS consistently outperforms baseline methods. Compared to Vanilla 3DGS, it achieves an average improvement of 23.8\% in SSIM, 1.59 dB in PSNR, and a 13.3\% reduction in LPIPS, with even greater improvements over Mirror-3DGS, averaging 36.6\% in SSIM, 2.32 dB in PSNR, and a 26.6\% decrease in LPIPS. Despite challenges like object structure and mirror equation errors, it remains overall more effective. Using mirror reflections, it enhances reconstruction accuracy, particularly in detail supplementation, while maintaining high-quality rendering with competitive training speed.

\subsection{Ablation Study}
\begin{figure}[t]
    \centering
    \includegraphics[width=0.7\linewidth]{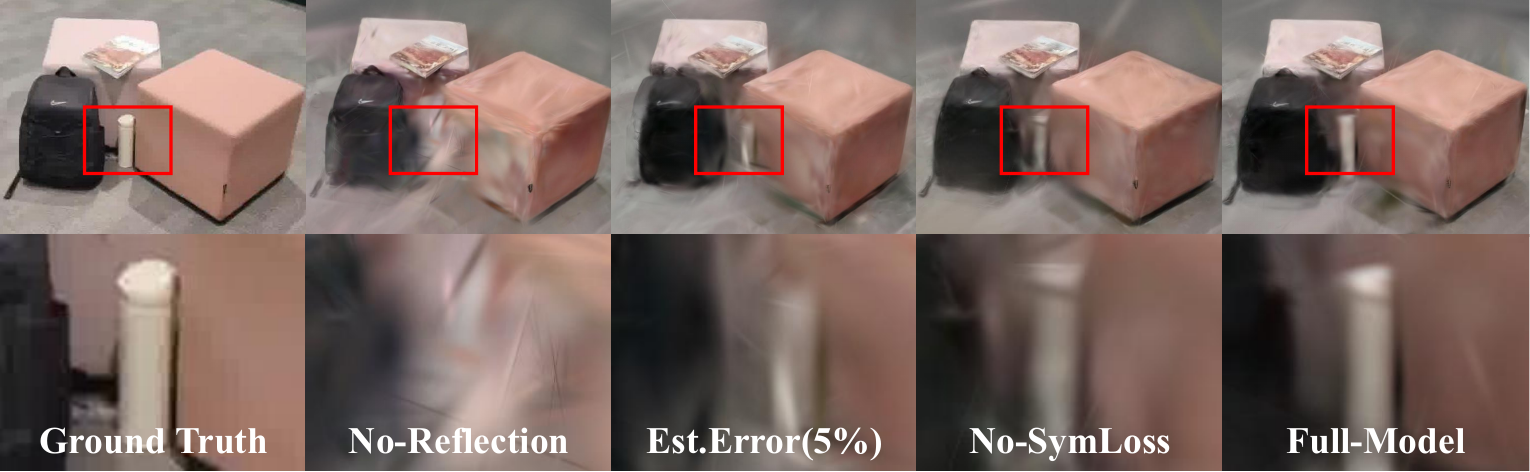}
    \caption{\textbf{Ablation Study Visualization (‘Bag’ Scene Example).} The figure presents full-scene and detail view renderings across three ablation experiments. Removing mirror reflection loses supplementary details, mirror equation errors cause blurring and ghosting, and omitting symmetry consistency loss results in inaccurate renderings and unclear boundaries. These components collectively enhance reconstruction quality.}
    \label{fig:ablation}
\end{figure}

We perform ablation experiments to evaluate the impact of key components: Mirror Information SupplementatioCn, Mirror Equation Estimation, and Symmetry Consistency Loss. Table~\ref{tab:ablation_comparison} summarizes the quantitative results, and Figure~\ref{fig:ablation} presents the visual comparisons.

\noindent\textbf{Mirror Information Supplementation.}
A white overlay mask was applied to the mirror regions to evaluate the impact of reflection information on reconstruction quality. The results show that covering the mirror reflections leads to a loss of supplementary information, preventing the model from recovering details, particularly those from occluded or viewpoint-blind areas. This highlights the critical role of mirror reflections in enhancing reconstruction quality.

\noindent\textbf{Mirror Equation Estimation.}
An error term was introduced to the mirror equation, causing a deviation 5\% from the parameters originally fitted. The results show visible ghosting in rendered images and degraded metrics, demonstrating precise mirror equation estimation is essential for maintaining symmetry between real and virtual Gaussians, forming a foundation for symmetry and joint optimization.

\noindent\textbf{Symmetry Consistency Loss.}
The symmetry consistency loss was removed by setting its hyperparameter to 0. This led to misalignment between real and mirrored Gaussians, causing unclear boundaries and inaccuracy rendering. The results confirm the necessity of symmetry constraints in preserving structural coherence and improving reconstruction accuracy.

\begin{table*}[t]
\centering
\resizebox{\textwidth}{!}{ 
\begin{tabular}{l|ccc|ccc|ccc|ccc|ccc|ccc}
\toprule
\textbf{Scene}                 & \multicolumn{3}{c|}{\textit{Cup}}              & \multicolumn{3}{c|}{\textit{Bag}}              & \multicolumn{3}{c|}{\textit{Duck}}             & \multicolumn{3}{c|}{\textit{Book}}            & \multicolumn{3}{c|}{\textit{Fruit}}       & \multicolumn{3}{c}{\textit{Average}}     \\ 
\cmidrule(lr){2-4} \cmidrule(lr){5-7} \cmidrule(lr){8-10} \cmidrule(lr){11-13} \cmidrule(lr){14-16}  \cmidrule(lr){17-19}
\textbf{Metrics} & SSIM$\uparrow$   & PSNR$\uparrow$   & LPIPS$\downarrow$   & SSIM$\uparrow$   & PSNR$\uparrow$   & LPIPS$\downarrow$    & SSIM$\uparrow$   & PSNR$\uparrow$   & LPIPS$\downarrow$   & SSIM$\uparrow$   & PSNR$\uparrow$   & LPIPS$\downarrow$   & SSIM$\uparrow$   & PSNR$\uparrow$   & LPIPS$\downarrow$   & SSIM$\uparrow$   & PSNR$\uparrow$   & LPIPS$\downarrow$   \\
\midrule
No-MirrorReflection                   & 0.63    & 17.83   & \textbf{0.33}    & 0.81    & \textbf{18.47}   & 0.23    & 0.70    & 19.07   & \textbf{0.32}  & 0.46    & 13.11   & 0.49    & 0.79    &17.27   & 0.24   & 0.68   &17.15   &0.32  \\
MirrorEstimationError(5\%)       & 0.68    & 17.27   & 0.40    & 0.85    &18.28   & 0.23    & 0.77    & 19.76  &0.35  & 0.84    & 15.05   & 0.36    & 0.68    &13.43   & 0.32   & 0.76   &16.76   &0.33  \\
No-SymmetryConsistencyLoss               & 0.73    & 18.26   & 0.42    & 0.87    & 18.00   & 0.20     & 0.77    & 20.08   & 0.41   & 0.89    & 16.27   & 0.32      & 0.88    & 17.92   & 0.19   & 0.83   &18.11   &0.31  \\
\textbf{Full-Model}                  & \textbf{0.82} & \textbf{19.49} & 0.37    & \textbf{0.90} & 18.27 & \textbf{0.19}     & \textbf{0.78} & \textbf{20.15} & 0.34     & \textbf{0.91} & \textbf{17.11} & \textbf{0.28}   & \textbf{0.90} & \textbf{19.05} & \textbf{0.18}   & \textbf{0.86} & \textbf{18.81} & \textbf{0.27}   \\ 
\bottomrule
\end{tabular}
}
\caption{\textbf{Ablation Study Results for SSIM, PSNR, and LPIPS.} In the ablation study, we tested the contributions of mirror information supplementation and the symmetry consistency loss function to the overall method. The complete model with all components consistently achieved the best performance. The best values highlighted in bold.}
\label{tab:ablation_comparison}
\end{table*}

\section{Conclusion}
\label{conclusion}

In this work, We introduced \textbf{ReflectiveGS}, a novel approach that integrates mirror reflections into 3D scene reconstruction. Unlike existing methods that focus on rendering mirror surfaces, our method leverages mirror reflections to recover absent viewpoints, enhancing reconstruction accuracy and detail preservation. To support this task, we constructed \textbf{MirrorScene3D}, a dedicated dataset for mirror-augmented reconstruction, providing a benchmark for evaluating the effectiveness of mirror reflections as supplementary information. Experiments show that ReflectiveGS outperforms Vanilla 3DGS and Mirror-3DGS in rendering quality, achieving higher SSIM and PSNR with lower LPIPS while maintaining competitive training efficiency.

Despite these advancements, certain challenges remain. Mirror estimation errors can occur in complex or weakly reflective regions, affecting the model's performance and requiring further optimization. Additionally, scalability to large-scale real scenes requires further investigation. Future work will focus on improving mirror estimation robustness, optimizing computational efficiency, and extending the approach to broader scene categories to enhance its real-world applicability.

\bibliographystyle{plain} 
\bibliography{references} 

\end{document}